\documentclass[10pt,journal,final]{IEEEtran}
\ifCLASSINFOpdf
  \usepackage[pdftex]{graphicx}
  \usepackage{color}
  \usepackage{multirow}
  \usepackage{booktabs}
  
\else
\fi
\usepackage{gensymb}
\usepackage{amsfonts,amssymb}
\usepackage{amsmath}
\usepackage{url}
\usepackage{overpic}
\usepackage{bbm}

\usepackage{rotating}
\newcommand{\gr}{\rowcolor[gray]{.95}}
\usepackage{algorithmic}

\usepackage{algorithm}
\ifCLASSOPTIONcompsoc
  \usepackage[caption=false,font=normalsize,labelfont=sf,textfont=sf]{subfig}
\else
  \usepackage[caption=false,font=footnotesize]{subfig}
\fi
 \usepackage{cite}
 \usepackage{dblfloatfix}
 \usepackage{xcolor}
\usepackage{colortbl}

\usepackage{hyperref}

\usepackage{color}

\makeatletter
\let\myorg@bibitem\bibitem
\def\bibitem#1#2\par{%
  \@ifundefined{bibitem@#1}{%
    \myorg@bibitem{#1}#2\par
  }{%
    \begingroup
      \color{\csname bibitem@#1\endcsname}%
      \myorg@bibitem{#1}#2\par
    \endgroup
  }%
}

\makeatother

\begin{document}
%\title{High-Quality Angle Prediction for Oriented Object Detection}
%Estimating the Quality of Angle Prediction forOriented Object Detection
\title{S$^2$Mamba: A Spatial-spectral State Space Model for Hyperspectral Image Classification}
%\author{Guanchun Wang}%
\author{Guanchun~Wang,~Xiangrong~Zhang,~\IEEEmembership{Senior~Member,~IEEE},~Zelin~Peng,~Tianyang~Zhang,~and~Licheng~Jiao,~\IEEEmembership{Fellow,~IEEE} \thanks{This work was supported in part by the National Natural Science Foundation of China under Grant 62276197, Grant 62171332, and the Fundamental Research Funds for the Central Universities Grant ZYTS24002.
 \textit{(Corresponding author: Xiangrong Zhang.)}

Guanchun Wang, Xiangrong Zhang, Tianyang Zhang, and Licheng Jiao are with the School of Artificial Intelligence, Xidian University, Xi’an 710071, China (e-mail: xrzhang@mail.xidian.edu.cn). 

Zelin Peng is with MoE Key Lab of Artificial Intelligence, AI Institute, Shanghai Jiao Tong University, Shanghai, 200240, China.
}}%
\maketitle

\begin{abstract}
Land cover analysis using hyperspectral images (HSI) remains an open problem due to their low spatial resolution and complex spectral information. Recent studies are primarily dedicated to designing Transformer-based architectures for spatial-spectral long-range dependencies modeling, which is computationally expensive with quadratic complexity. Selective structured state space model (Mamba), which is efficient for modeling long-range dependencies with linear complexity, has recently shown promising progress. However, its potential in hyperspectral image processing that requires handling numerous spectral bands has not yet been explored. In this paper, we innovatively propose \textbf{S$^2$Mamba}, a spatial-spectral state space model for hyperspectral image classification, to excavate spatial-spectral contextual features, resulting in more efficient and accurate land cover analysis. In {S$^2$Mamba}, two selective structured state space models through different dimensions are designed for feature extraction, one for spatial, and the other for spectral, along with a spatial-spectral mixture gate for optimal fusion. More specifically, S$^2$Mamba first captures spatial contextual relations by interacting each pixel with its adjacent through a Patch Cross Scanning module and then explores semantic information from continuous spectral bands through a Bi-directional Spectral Scanning module. Considering the distinct expertise of the two attributes in homogenous and complicated texture scenes, we realize the Spatial-spectral Mixture Gate by a group of learnable matrices, allowing for the adaptive incorporation of representations learned across different dimensions. Extensive experiments conducted on HSI classification benchmarks demonstrate the superiority and prospect of S$^2$Mamba. The code will be made available at: https://github.com/PURE-melo/S2Mamba.
\end{abstract}
\begin{IEEEkeywords}
Hyperspectral image classification, spatial-spectral state space model, spatial-spectral feature fusion, linear complexity
\end{IEEEkeywords}
\section{Introduction}
\label{section1}
\IEEEPARstart{H}{yperspectral} images (HSI), consisting of numerous spectral bands, are capable of land cover analysis due to their rich material information \cite{HSI-survey2,HSI-survey4,HSI-survey1,SpectralGPT}, with extensive application in domains such as precision agriculture, mineral exploration, and environmental monitoring \cite{HSI-survey3,cast}. Therefore, there is a strong incentive to design a more effective and efficient model for hyperspectral image classification. Convolutional neural networks (CNNs) \cite{CNN_spatial_2,CNN_SS_1,CNN_SS_2,GNN_1,GNN_2} as a widespread paradigm have been widely studied in hyperspectral image classification. However, this paradigm is limited by its local receptive fields and is unable to comprehensively capture continuous spectral properties. Transformer architectures \cite{Spectralformer,SSFTT,morphformer,transformer_3,transformer_5,HiT} have been recently explored in hyperspectral image classification, which exhibits remarkable performance due to their ability to extract global contextual information on both spatial and spectral dimensions.

Despite its powerful representation ability, dealing HSI data with transformer-based models is computationally expensive, primarily due to the self-attention mechanism with quadratic computational complexity $\mathcal{O}\left(N^2\right)$ \cite{transformer,Vit}. As an efficient alternative to self-attention mechanisms, the selective structured state space model (Mamba) \cite{Mamba} has recently emerged as a powerful tool with linear complexity for modeling long-range dependency in sequence processing. Derived from this, a series of Mamba-based models \cite{Visionmamba,Vmamba,Efficientvmamba,Rsmamba,Umamba,denseRSmamba,Segmamba,Videomamba,Pointmamba} have been explored for various computer vision tasks, such as image classification, semantic segmentation, and others. However, most of these models are only applied in natural image processing, leaving a blank in hyperspectral image classification due to the difficulty in handling complicated spatial-spectral information.

Drawing inspiration from the success of state space models, this paper seeks to explore their potential for HSI processing. To this end, we propose a spatial-spectral state space model, termed \textit{\textbf{S$^2$Mamba}}, to jointly excavate long-range spatial relations and continuous spectral features for hyperspectral image classification. Our S$^2$Mamba comprises \textit{\textbf{Patch Cross Scanning}} and \textit{\textbf{Bi-directional Spectral Scanning}} modules for capturing spatial and spectral information, respectively, and merging them with a \textit{\textbf{Spatial-spectral Mixture Gate}}. Specifically, we build a patch cross scanning mechanism to capture the contextual relations between adjacent pixels, wherein the patch data is first flattened into pixel sequences through different route generation ways and a selective structured state space model is then applied to these sequences for capturing contextual features. Considering the rich knowledge within continuous spectral bands, we design an extra scanning module on the spectral dimension for retrieving semantic properties in HSI data by a bi-directional interaction between each band.

\begin{figure*}[t]
	\centering
        % \vspace{-8pt}
	\begin{overpic}[width=0.97\linewidth]{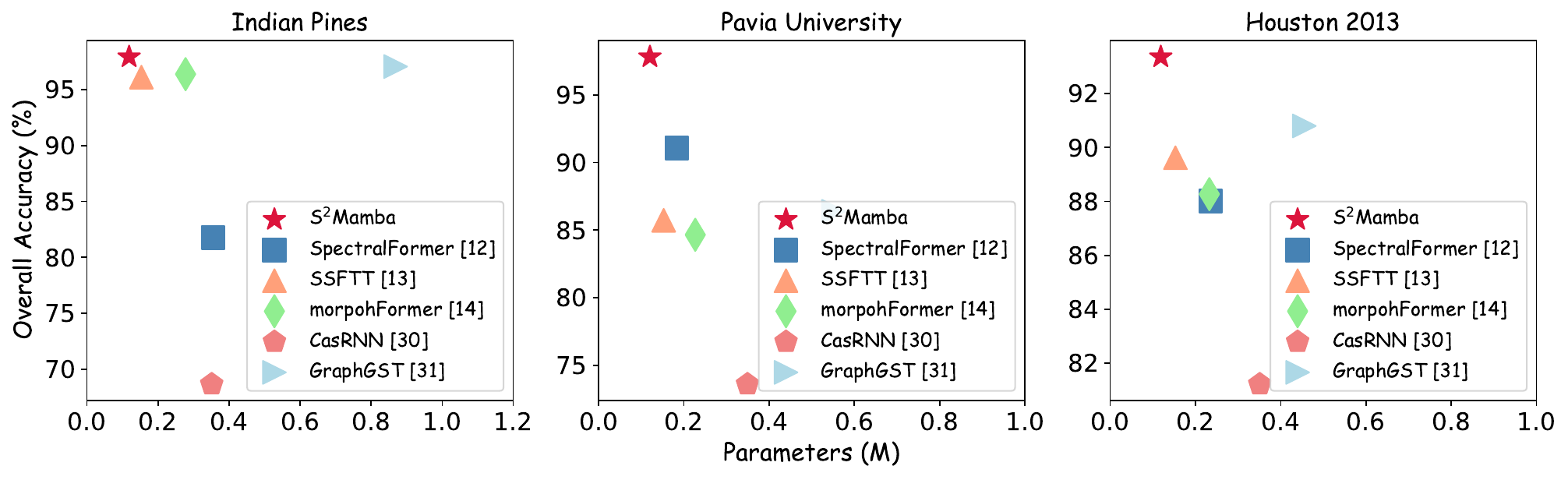}
	\end{overpic}
        % \vspace{-4pt}
	\caption{Overall Accuracy (\%) and Parameters (M) comparison on Indian Pines, Pavia University, Houston 2013 dataset. Our proposed S$^2$Mamba achieves optimal results in terms of overall accuracy and parameters compared to existing methods \cite{Spectralformer,SSFTT,morphformer,RNN_3,GraphGST}.}
    \label{fig:intro}
    % \vspace{-18pt}
\end{figure*}
The remaining dilemma lies in optimally merging spatial and spectral attributes of HSI data. We observe that spectral information exhibits a more significant role for uniform land cover regions than those regions with complex textures. This is due to the paucity of spatial cues within homogenous regions, necessitating the reliance on spectral information for determining land cover categories. In contrast, spatial information can offer a considerable prior in dealing with regions containing complex textures, thereby enhancing classification performance. To address this, we here propose a spatial-spectral mixture gate to dynamically merge the above features pixel by pixel, wherein each spatial location is assigned a group of learnable weights to determine the ratio of various features. This gating mechanism, promoting the competition of the two features, facilitates the integration of spatial-spectral attributes by truncating those redundant ones, greatly boosting the accuracy.

Through experimental assessment, we verify the effectiveness of S$^2$Mamba on the three public hyperspectral image classification datasets. Fig. \ref{fig:intro} illustrates that our S$^2$Mamba improves the previous SOTA method by {{0.86\%}}, {{6.74\%}}, and {{2.56\%}} in terms of overall accuracy on Indian Pines, Pavia University, and Houston 2013 datasets, respectively. Meanwhile, it outperforms Transformer-based models with {the fewest parameters (about 0.12M)} and a linear complexity $\mathcal{O}\left(kN\right)$, where $k \ll N$.

The contribution of our work can be summarized as follows:

1) A novel spatial-spectral state space model, termed S$^2$Mamba, is proposed for hyperspectral image classification, which is the innovative exploration to introduce Mamba into hyperspectral classification. Extensive experiments demonstrate that S$^2$Mamba can achieve state-of-the-art with the fewest parameters.

2) A patch cross scanning module (PCS) is designed to excavate spatial contextual relations between adjacent pixels by flattening HSI data into multiple pixel sequences and scanning them with a selective structured state space model, thereby facilitating land cover discrimination. 

3) Considering the rich prior within spectral, a bi-directional spectral scanning module (BSS) is proposed to fully explore the spectral discriminative properties by scanning continuous bands with a selective structured state space model, enhancing the classification capability of the model.

4) To optimally integrate spatial and spectral attributes of HSI, a spatial-spectral mixture gate is proposed, which controls the domination of the two features by adaptive feature competition, further ameliorating the model performance.

The remainder of this paper is organized as follows. In Section II, we briefly review related works on hyperspectral image classification and state space models. In Section III, we describe the details of the proposed hyperspectral image classification framework S$^2$Mamba. We introduce the dataset and evaluation metrics in Section IV, and show the experiment settings and results in Section V. Finally, Section VI concludes this paper.
\section{Related Work}

\subsection{Hyperspectral Image Classification}
Existing methods typically solve the hyperspectral image classification problem by leveraging well-crafted deep neural networks, which can be primarily divided into several categories: convolutional neural network-based \cite{CNN_SS_1,CNN_SS_2,CNN_SS_3,CNN_SS_4,CNN_SS_5,CNN_SS_6,CNN_spatial_1,CNN_spatial_2,CNN_spectral_1,CNN_spectral_2,GNN_1,GNN_2,GNN_3,GNN_4,FPGA,SPRN}, recurrent neural network-based \cite{RNN_1,RNN_2,RNN_3}, and Transformer-based \cite{transformer_1,SSFTT,transformer_3,transformer_5,Spectralformer,morphformer,HiT,mix_2,mix_3,GraphGST,multiscan}.

\subsubsection{Traditional Models} CNNs are widely applied in hyperspectral classification tasks, an excellent feature extraction architecture that captures spatial and local semantic information. Recent works have explored CNNs to individually extract spatial \cite{CNN_spatial_1,CNN_spatial_2} and spectral features \cite{CNN_spectral_1,CNN_spectral_2} from hyperspectral remote sensing images, or to learn spatial-spectral joint representation \cite{CNN_SS_1,CNN_SS_2,CNN_SS_3,CNN_SS_4,CNN_SS_5,CNN_SS_6}, achieving remarkable progress. Considering that CNN structures are incapable of modeling irregular data, some research introduced graph convolutional networks (GCNs) \cite{GNN_1,GNN_2,GNN_3,GNN_4} to mine the potential spatial semantic information of HSI data. Another group of studies serves the spectral information across different bands of HSI data as continuous sequences and employs recurrent neural networks (RNNs) \cite{RNN_1,RNN_2,RNN_3} to extract spectral features for classification. However, these methods often struggle to extract global spectral information due to their limited capability in long-range dependency.

\begin{figure*}[t]
	\centering
	\begin{overpic}[width=\linewidth]{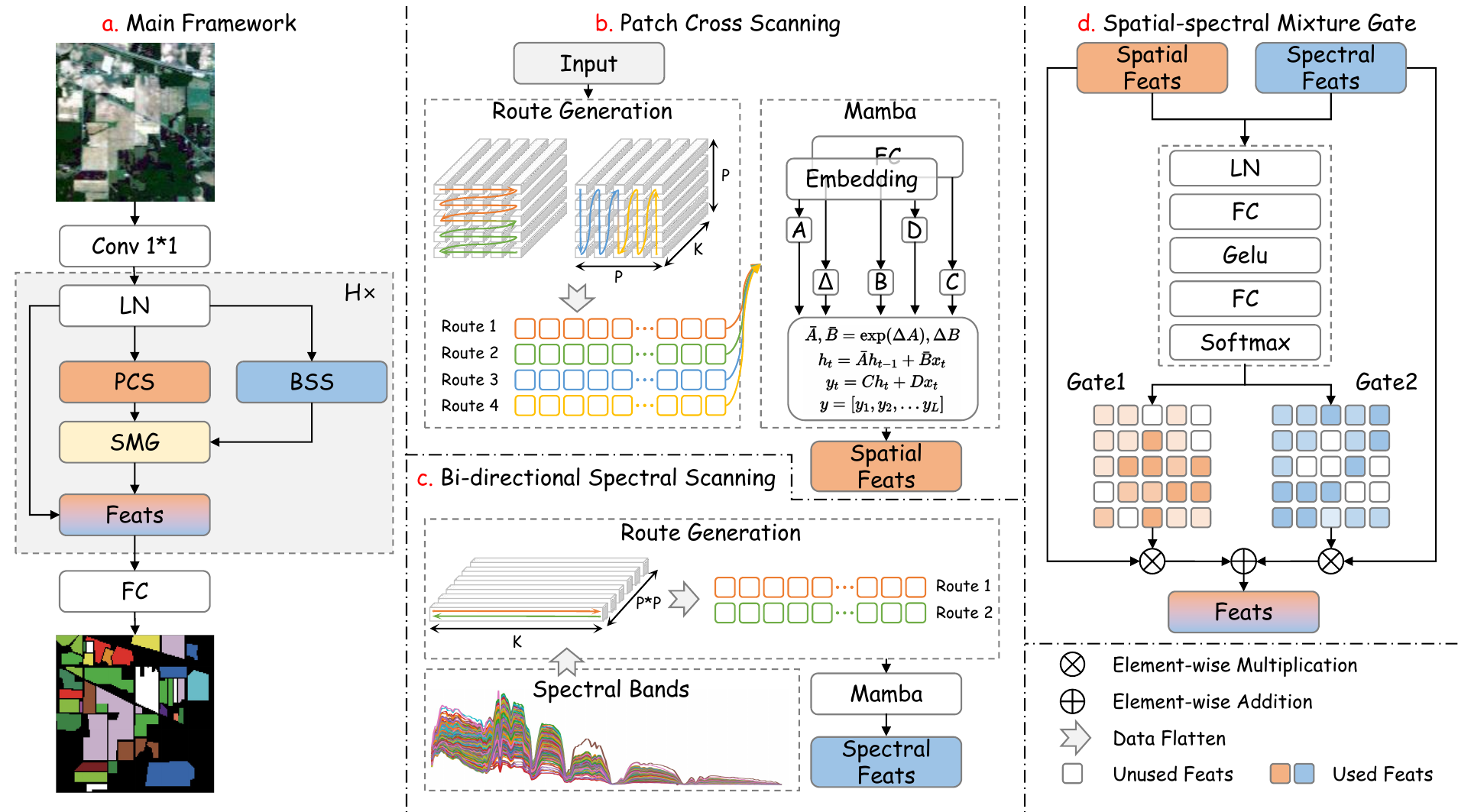}
	\end{overpic}
	\caption{Illustration of our proposed S$^2$Mamba, consisting of a patch cross scanning (PCS) mechanism, a bi-directional spectral scanning (BSS) mechanism, and a spatial-spectral mixture gate (SMG).}
    \label{fig:overall}
    % \vspace{-15pt}
\end{figure*}
\subsubsection{Transformer-based Models} Transformer is a powerful architecture consisting of multiple self-attention mechanisms to extract global contextual information, and currently, they have been explored for hyperspectral image classification. Most of these methods \cite{transformer_1,SSFTT,transformer_3,transformer_5,Spectralformer,morphformer,HiT,mix_2,mix_3,GraphGST,multiscan} attempt to learn the global sequential information on both spatial and spectral dimensions. Spectralformer \cite{Spectralformer} is the first one in its kind to introduce the Transformer architecture into hyperspectral image classification, which jointly captures the local and global information by grouping the adjacent bands. Beyond pure Transformer structures, some methods adopt hybrid networks to acquire spatial-spectral features. SSFTT \cite{SSFTT} utilizes convolutional layers to describe the low-level features and integrate them through Transformer layers. morphFormer \cite{morphformer} uses morphological convolution layers to learn both spatial and spectral representations and merges them by applying Transformer layers. Besides, other works \cite{multiscan} incorporate RNNs with Transformers to collaboratively extract continuous spectral features and spatial context features, enhancing the classification performance. 

Unlike previous methods that capture long-range dependencies by compute-intensive Transformer structures, we innovatively explore a fully sequential architecture based on selective structured state space models for efficient global spatial-spectral feature extraction.

\subsection{State Space Models}
Most recently, state space models (SSMs) \cite{hippo,S4,SSM,SSM2,SSM3}, especially the structured state space models (S4) \cite{S4} have shown promising progress in sequence analysis, which are capable of long-range sequence modeling with linear computational complexity. By introducing a selective mechanism into \cite{S4}, Mamba \cite{Mamba} further optimizes its context compression ability and exhibits superior performance to Transformers. Considering its outstanding performance in sequence data processing, many works have explored the potential of Mamba \cite{Mamba} in computer vision and achieved promising advancements. In particular, visual state space model (Vmamba) \cite{Vmamba} and vision Mamba (Vim) \cite{Visionmamba} have recently emerged as powerful tools for various computer vision tasks due to their efficiency in modeling long-range dependencies. On the basis of them, a series of visual state space models have been proposed, such as medical image analysis \cite{Umamba,Segmamba,VMUnet,Swinumamba}, video understanding \cite{Videomamba,videomamba2}, and others \cite{denseRSmamba,Rsmamba,Panmamba,Pointmamba,Changemamba,Efficientvmamba,RS3Mamba}. However, most of these methods are only applied in RGB images, leaving a blank in hyperspectral image classification that requires handling complex spectral information. Thus, we propose a Mamba-based architecture for hyperspectral image classification, fully exploiting spatial-spectral features by state space models.

\section{Proposed Method}

\subsection{Preliminaries}

\subsubsection{State Space Models}
State space models are foundational statistical models that describe the dynamic behavior of systems and are widely applied in domains such as time series analysis and control systems. In SSMs, the continuous evolution of systems is calculated through a set of ordinary differential equations (ODEs), which map an input signal into a latent space and decode it into an output sequence. This operation can be defined as:

\begin{equation}
\begin{aligned}
& h^{\prime}(t)=\mathbf{A} h(t)+\mathbf{B} x(t) \\
& y(t)=\mathbf{C} h(t)+\mathbf{D} x(t)
\end{aligned},
\end{equation}
where $h(t) \in \mathbb{R}^{N}$, $x(t) \in \mathbb{R}^{L}$, and $y(t) \in \mathbb{R}^{L}$ indicate latent state, input signal, and output signal, respectively. $h^{\prime}(t) \in \mathbb{R}^{N}$ refers to the time derivative of $h(t)$. $N$ and $L$ represent the dimensions of the latent space and sequences, respectively. Additionally, $\mathbf{A} \in \mathbb{R}^{N \times N}$ is the state transition matrix. $\mathbf{B} \in \mathbb{R}^{N\times L}$ and $\mathbf{C} \in \mathbb{R}^{L \times N}$ are projection matrices. $\mathbf{D}$ is usually served as a residual connected operation and thus discarded in equations. To facilitate continuous state space models with discrete sequences, Mamba \cite{Mamba} employs a zero-order hold technique to discretize the ordinary differential equations:
\begin{equation}
\begin{aligned}
& \overline{\mathbf{A}}=\exp ({\Delta \mathbf{A}}) \\
& \overline{\mathbf{B}}=({\Delta} \mathbf{A})^{-1}(\exp (\Delta \mathbf{A})-\mathbf{I}) \cdot \Delta \mathbf{B}
\end{aligned},
\end{equation}
where $\overline{\mathbf{A}}$ and $\overline{\mathbf{B}}$ represent the discretized forms of the parameters $\mathbf{A}$ and $\mathbf{B}$, respectively, through a discretization step size $\Delta$. As stated in \cite{Mamba}, the projection matrix $\overline{\mathbf{B}}$ can be approximated using the first-order Taylor series:
\begin{equation}
    \overline{\mathbf{B}} =(\exp (\mathbf{A}) - I) \mathbf{A}^{-1} \mathbf{B} \approx ({\Delta} \mathbf{A})({\Delta} \mathbf{A})^{-1}{\Delta}\mathbf{B} = \mathrm{\Delta}\mathbf{B},
\end{equation}

After discretization, the ODEs of SSMs can be represented as follows: 
\begin{equation}
\label{eq4}
\begin{aligned}
h_t & =\overline{\mathbf{A}} h_{t-1}+\overline{\mathbf{B}} x_t \\
y_t & =\mathbf{C} h_t 
\end{aligned},
\end{equation}

\subsubsection{Selective Scan Mechanism.}
Traditional SSMs are linear time-invariant, i.e., the projection matrices do not vary with input signals, resulting in non-selective attention on each sequence unit. To alleviate this, Mamba \cite{Mamba} modifies the parameter matrices to be input-dependent, i.e., $\mathbf{B} \in \mathbb{R}^{B \times L \times N}$, $\mathbf{C} \in \mathbb{R}^{B \times L \times N}$ and $\Delta \in \mathbb{R}^{B \times L \times D}$ are calculated based on $\mathbf{X} \in \mathbb{R}^{B \times L \times D}$, further improving the ability for handling complex sequences by transforming the SSMs into linear time-varying systems.

\subsection{S$^2$Mamba}
\subsubsection{\textbf{Overview}}
Fig. \ref{fig:overall} (a) illustrates the overall network architecture of our proposed S$^2$Mamba framework for hyperspectral image classification. Given the input data $\mathbf{X} \in \mathbb{R}^{P \times P \times K}$, where $P$ and $K$ denote the patch size and spectral band number of the data cube, respectively, which is first fed into a 1 $\times$ 1 convolutional layer for embedding. After this process, we can obtain the encoded feature $\mathbf{E} \in \mathbb{R}^{P \times P \times D}$, where $D$ denotes the latent dimension. Then, we feed $\mathbf{E} \in \mathbb{R}^{P \times P \times D}$ into patch cross scanning and bi-directional spectral scanning modules to capture spatial and spectral features, respectively. Next, the above terms are fused by a spatial-spectral mixture gate for fusing. The structure of the spatial-spectral mixture gate is detailed in Fig. \ref{fig:overall} (d), which simply comprises two fully connected layers with activation functions. Finally, the fused spatial-spectral features are fed into a fully connected layer for category prediction.

Notably, our S$^2$Mamba is extremely lightweight but efficient, where the number of layers $H$ is only set to 1, and the latent dimension $D$ is set to 64. Moreover, unlike recent advanced approaches that utilize Transformer-based networks with quadratic complexity to capture global spatial-spectral features, our S$^2$Mamba is a powerful spatial-spectral information extraction network with linear computational complexity.

\subsubsection{\textbf{Patch Cross Scanning Mechanism}}
To facilitate the selective scanning mechanism with HSI inputs, we first extend the vanilla selective scan mechanism \cite{Mamba,Vmamba} to the patch-level HSI data and design a patch cross scanning mechanism, which captures spatial contextual relations by interacting each pixel with its adjacent through a state space model. Given the input of HSI patch $\mathbf{X} \in \mathbb{R}^{P \times P \times K}$, where $P$ and $K$ denote the patch size and spectral band number of the data cube, respectively, we perform a pixel-by-pixel scanning strategy on four different routes. As illustrated in Fig. \ref{fig:overall}, each route is generated from different directions, such as top to bottom, left to right, and vice versa.

More specifically, we first flatten the patch data into one-dimensional sequences with the preset routes, and then recurrently calculate each item of sequences $\mathcal{X}_{\text{seq}}=\{[\mathbf{x}^i_0, \mathbf{x}^i_1,\ldots, \mathbf{x}^i_{P \times P}]\ |\mathbf{x}^i_j \in \mathbb{R}^{1 \times K}, i\in \{0,1,2,3\}\}$ by using rewritten Eq.~(\ref{eq4}):
\begin{equation}
\begin{aligned}
\mathbf{h}^i_j & =\overline{\mathbf{A}}_{\text{spa}} \mathbf{h}^i_{j-1}+\overline{\mathbf{B}}_{\text{spa}} \mathbf{x}^i_j \\
\mathbf{y}^i_j & =\mathbf{C}_{\text{spa}} \mathbf{h}^i_j + \mathbf{x}^i_j 
\end{aligned},
\end{equation}
where $\overline{\mathbf{A}}_{\text{spa}}$, $\overline{\mathbf{B}}_{\text{spa}}$ and ${\mathbf{C}}_{\text{ spa}}$ represent the trainable parameters in PCS. After scanning, we can obtain a set of output sequences:
\begin{equation}
\mathcal{Y}_{\text{seq}}=\{[\mathbf{y}^i_0, \mathbf{y}^i_1,\ldots, \mathbf{y}^i_{P \times P}]\ | \mathbf{y}^i_j \in \mathbb{R}^{1 \times K}, i\in \{0,1,2,3\}\},
\end{equation}

Next, the output sequences obtained from different scanning routes are fused according to the operations in Fig. \ref{fig:overall}, such as flipping or transposing sequences. As a result, each element in the output sequence $\mathbf{Y} \in \mathbb{R}^{P \times P \times K}$ can integrate influences with its adjacent regions from different directions.

\subsubsection{\textbf{Bi-directional Spectral Scanning Mechanism}}
Although the above scanning mechanism involves the spatial contextual information within the data cube, it lacks consideration for continuous spectral band information inherent in HSI data. One naive approach to address this issue is to scan the data cube band by band, capturing the semantic cues from rich spectral bands. However, due to the unidirectional information induction attribute of state space models, the spectral scanning mechanism conducted in a single direction may fail to adequately capture the contextual information between spectral bands, leading to limited spectral utilization.

To this end, we further design a bi-directional spectral scanning mechanism, analyzing the varying tendency of the continuous spectrum from multiple directions by scanning spectral dimension band by band. We first flatten the HSI patch along the spatial dimensions to acquire the data matrix $\mathbf{S} \in \mathbb{R}^{K \times  P^2}$, and then recurrently calculate each item of sequences $\mathcal{S}_{seq}=\{[\mathbf{s}^i_0, \mathbf{s}^i_1,\ldots, \mathbf{s}^i_{K}]\ |\mathbf{s}^i_j \in \mathbb{R}^{1 \times P^2}, i\in \{0,1\}\}$ by following operation:
\begin{equation}
\begin{aligned}
\tilde{\mathbf{h}}^i_j & =\overline{\mathbf{A}}_{\text{spe}} \tilde{\mathbf{h}}^i_{j-1}+\overline{\mathbf{B}}_{\text{spe}} \mathbf{s}^i_j \\
\mathbf{p}^i_j & =\mathbf{C}_{\text{spe}} \tilde{\mathbf{h}}^i_j + \mathbf{s}^i_j 
\end{aligned},
\end{equation}
where $\mathbf{p}^i_j$ denotes the $j$th element in the $i$th order of output sequences. $\overline{\mathbf{A}}_{\text{spe}}$, $\overline{\mathbf{B}}_{\text{spe}}$ and ${\mathbf{C}}_{\text{ spe}}$ represent the trainable parameters in BSS. After scanning, we can obtain a set of output sequences:

\begin{equation}
\mathcal{P}_{\text{seq}}=\{[\mathbf{p}^i_0, \mathbf{p}^i_1,\ldots, \mathbf{p}^i_{K}]\ | \mathbf{p}^i_j \in \mathbb{R}^{1 \times P^2}, i\in \{0,1\}\},
\end{equation}

Next, the output sequences $\mathbf{P} \in \mathbb{R}^{P \times P \times K}$ are fused from different routes, which integrate influences from adjacent bands into each spectral band, further boosting its discriminative ability.

\begin{table}[!t]
\centering
\caption{Land-cover categories of the Indian Pines dataset and its corresponding training and testing samples.}
\begin{tabular}{c||ccc|c}
\toprule[1.5pt]
No. & Category & Training & Testing & Color \\
\hline \hline
1 & Corn Notill & 50 & 1384 & \cellcolor[RGB]{83,171,72} \\
2 & Corn Mintill & 50 & 784 & \cellcolor[RGB]{137,186,67} \\
3 & Corn & 50 & 184 & \cellcolor[RGB]{66,132,91} \\
4 & Grass Pasture & 50 & 447 & \cellcolor[RGB]{60,131,69} \\
5 & Grass Trees & 50 & 697 & \cellcolor[RGB]{144,82,54} \\
6 & Hay Windrowed & 50 & 439 & \cellcolor[RGB]{105,188,200} \\
7 & Soybean Notill & 50 & 918 & \cellcolor[RGB]{255,255,255} \\
8 & Soybean Mintill & 50 & 2418 & \cellcolor[RGB]{199,176,201} \\
9 & Soybean Clean & 50 & 564 & \cellcolor[RGB]{218,51,44} \\
10 & Wheat & 50 & 162 & \cellcolor[RGB]{119,35,36} \\
11 & Woods & 50 & 1244 & \cellcolor[RGB]{55,101,166} \\
12 & Buildings Grass Trees Drives & 50 & 330 & \cellcolor[RGB]{224,219,84} \\
13 & Stone Steel Towers & 50 & 45 & \cellcolor[RGB]{217,142,52} \\
14 & Alfalfa & 15 & 39 & \cellcolor[RGB]{84,48,126} \\
15 & Grass Pasture Mowed & 15 & 11 & \cellcolor[RGB]{227,119,91} \\
16 & Oats & 15 & 5 & \cellcolor[RGB]{157,87,150} \\
\hline \hline
 & Total & 695 & 9671 & \\
\bottomrule[1.5pt]
\end{tabular}
\label{data1}
\end{table}

\begin{table}[!t]
\centering
\caption{Land-cover categories of the Pavia University dataset and its corresponding training and testing samples.}
\begin{tabular}{c||ccc|c}
\toprule[1.5pt]
No. & Category & Training & Testing & Color \\
\hline \hline
1 & Asphalt & 548 & 6304 & \cellcolor[RGB]{83,171,72} \\
2 & Meadows & 540 & 18146 & \cellcolor[RGB]{66,132,91} \\
3 & Gravel & 392 & 1815 & \cellcolor[RGB]{144,82,54} \\
4 & Trees & 524 & 2912 & \cellcolor[RGB]{255,255,255} \\
5 & Metal Sheets & 265 & 1113 & \cellcolor[RGB]{218,51,44} \\
6 & Bare Soil & 532 & 4572 & \cellcolor[RGB]{55,101,166}  \\
7 & Bitumen & 375 & 981 & \cellcolor[RGB]{217,142,52} \\
8 & Bricks & 514 & 3364 & \cellcolor[RGB]{227,119,91} \\
9 & Shadows & 231 & 795 & \cellcolor[RGB]{157,87,150} \\
\hline \hline
 & Total & 3921 & 40002 & \\
\bottomrule[1.5pt]
\end{tabular}
\label{data2}
\end{table}

\subsubsection{\textbf{Spatial-spectral Mixture Gate}}
After acquiring the spatial and spectral information of HSI through the two scanning modules, calculating the optimal mixture representation becomes a key challenge. As discussed in Section \ref{section1}, the effectiveness of spatial and spectral features in HSI classification differs across different scenarios, thereby their direct merging without prior may lead to contradictions. 

Specifically, we note that spectral information acts more prominence in the classification of uniform regions as opposed to those characterized by complicated textures. This is attributable to the scarcity of spatial cues within homogenous regions, which in turn emphasizes the importance of spectral information for differentiating these land cover. In this case, PCS may contribute to misleading directions as a consequence of the redundant features. Inversely, it offers considerable prior in dealing with regions characterized by complex textures, thereby enhancing the richness of discriminative representation. 

\begin{table}[!t]
\centering
\caption{Land-cover categories of the Houston 2013 dataset and its corresponding training and testing samples.}
\begin{tabular}{c||ccc|c}
\toprule[1.5pt]
No. & Category & Training & Testing & Color \\
\hline \hline
1 & Healthy Grass & 198 & 1053 & \cellcolor[RGB]{83,171,72} \\
2 & Stressed Grass & 190 & 1064 & \cellcolor[RGB]{137,186,67} \\
3 & Synthetic Grass & 192 & 505 & \cellcolor[RGB]{66,132,91} \\
4 & Tree & 188 & 1056 & \cellcolor[RGB]{60,131,69} \\
5 & Soil & 186 & 1056 & \cellcolor[RGB]{144,82,54} \\
6 & Water & 182 & 143 & \cellcolor[RGB]{105,188,200} \\
7 & Residential & 196 & 1072 & \cellcolor[RGB]{255,255,255} \\
8 & Commercial & 191 & 1053 & \cellcolor[RGB]{218,51,44} \\
9 & Road & 193 & 1059 & \cellcolor[RGB]{119,35,36} \\
10 & Highway & 191 & 1036 & \cellcolor[RGB]{55,101,166} \\
11 & Railway & 181 & 1054 & \cellcolor[RGB]{224,219,84} \\
12 & Parking Lot1 & 192 & 1041 & \cellcolor[RGB]{ 217,142,52} \\
13 & Parking Lot2 & 184 & 285 & \cellcolor[RGB]{ 84,48,126} \\
14 & Tennis Court & 181 & 247 & \cellcolor[RGB]{ 227,119,91} \\
15 & Running Track & 187 & 473 & \cellcolor[RGB]{157,87,150} \\
\hline \hline
 & Total & 2832 & 12197 & \\
\bottomrule[1.5pt]
\end{tabular}
\label{data3}
\end{table}

\begin{table*}[!t]
\centering
\caption{Comparison with other state-of-the-art methods in terms of (OA\%/AA\%/$\kappa$) on the Indian Pines test set. The best is in bold.}
\vspace{-4pt}
\resizebox{0.99\textwidth}{!}{
\begin{tabular}{c||ccc|cc|ccccc|c}
\toprule[1.5pt] \multirow{2}{*}{Category} & \multicolumn{3}{c|}{CNNs} & \multicolumn{2}{c|}{RNN} & \multicolumn{5}{c|}{Transformers} & \multicolumn{1}{c}{Mamba}\\
\cline{2-12}  
% \specialrule{0em}{1pt}{1pt}
& 1-D CNN & 2-D CNN & miniGCN & RNN & CasRNN& ViT & SpectralFormer & morphFormer & SSFTT& GraphGST & \bf S$^2$Mamba\\
\hline \hline
1 & 47.83 & 65.90 &  72.54 & 69.00 &61.78 &  53.25  & 70.52& 93.14&91.18 &\bf 95.81&94.44 \\
2 & 42.35 & 76.66 & 55.99 & 58.93  & 57.78& 66.20 &  81.89&97.70 &98.72 &98.85&\bf100.00 \\
3 & 60.87 & 92.39 & 92.93 & 77.17& 75.00& 86.41  & 91.30&100.00&100.00 &100.00&\bf100.00 \\
4 & 89.49 & 93.96 & 92.62 & 82.33 & 90.16& 89.71 &  95.53& 96.87&96.19 &97.09&\bf98.43 \\
5 & 92.40 & 87.23 &  94.98& 67.72  & 81.35& 87.66 & 85.51& 99.86&100.00 &98.57&\bf100.00 \\
6 & 97.04 & 97.27 & 98.63 & 89.07 & 87.70& 89.98 &  99.32& 99.77&100.00 &99.54&\bf100.00 \\
7 & 59.69 & 77.23 & 64.71& 69.06  & 79.08& 72.22 & 81.81& 86.71&95.10 &97.93&\bf98.47 \\
8 & 65.38 & 57.03  & 68.78 & 63.56&56.29 & 66.00  &  75.48&97.93&94.44 &94.50&\bf98.10 \\
9 &  93.44 & 72.87 & 69.33 & 65.07 & 60.11& 57.09  & 73.76& 94.33&90.96 &95.04&\bf95.04 \\
10 & 99.38 &  100.00  & 98.77& 95.06 & 95.06& 97.53 & 98.77&100.00&100.00 &98.76&\bf100.00 \\
11 & 84.00 & 92.85 & 87.78& 88.67  &82.23 & 87.62 &  93.17&99.12&99.35 &\bf99.59&97.67 \\
12 & 86.06 &  88.18 & 50.00 & 50.00 &46.97 & 63.94 & 78.48&99.09&100.00 &98.48&\bf100.00 \\
13 & 91.11 &  100.00 &  100.00& 97.78  & 97.78& 95.56 &  100.00& 100.00&100.00 &100.00&\bf100.00 \\
14 & 84.62 & 84.62 & 48.72& 66.67  &56.41 & 79.49 & 79.49 &92.31 &100.00 &100.00&\bf100.00 \\
15 &  100.00 &  100.00 & 72.73& 81.82  &81.82 & 90.91 & 100.00&100.00 &100.00 &100.00&\bf100.00 \\
16 & 80.00 &  100.00 & 80.00&  100.00  & 100.00& 80.00  &  100.00& 100.00&100.00 &100.00&\bf100.00 \\
\hline \hline
\gr OA (\%) & 70.43 & 75.89 & 75.11 & 70.66 & 68.65& 71.86  &  81.76 &96.38 &96.11 &97.06 &\bf97.92\\
\gr AA (\%) & 79.60 & 86.64 & 78.03 & 76.37 &75.59 & 78.97 &  87.81 & 97.30&97.92 &98.39 &\bf98.88\\
\gr $\kappa$  & 0.6642 & 0.7281 & 0.7164 & 0.6673 & 0.6464& 0.6804 & 0.7919 & 0.9584&0.9555 &0.9664 &\bf0.9761\\
\bottomrule[1.5pt]
\end{tabular}}
\label{IP-result}
\end{table*}

\begin{table*}[!t]
\centering
\caption{Comparison with other state-of-the-art methods in terms of (OA\%/AA\%/$\kappa$) on the Pavia University test set. The best is in bold.}
% \vspace{-4pt}
\resizebox{0.99\textwidth}{!}{
\begin{tabular}{c||ccc|cc|ccccc|c}
\toprule[1.5pt] \multirow{2}{*}{Category} & \multicolumn{3}{c|}{CNNs} & \multicolumn{2}{c|}{RNN} & \multicolumn{5}{c|}{Transformers} & \multicolumn{1}{c}{Mamba}\\
\cline{2-12}  
% \specialrule{0em}{1pt}{1pt}
& 1-D CNN & 2-D CNN & miniGCN & RNN & CasRNN& ViT & SpectralFormer & morphFormer & SSFTT& GraphGST & \bf S$^2$Mamba\\
\hline \hline
1 & 88.90 & 80.98 &   \bf96.35 &84.01 & 77.62& 71.51 & 82.73& 89.90&87.64&84.99 & 96.24 \\
2 & 58.81 & 81.70 &  89.43& 66.95 & 63.41& 76.82 & 94.03& 75.26& 76.60& 82.43& \bf98.75 \\
3 & 73.11 & 67.99 &  \bf87.01 & 58.46 &57.30& 46.39 & 73.66&85.90 & 85.56& 79.94&85.95  \\
4 & 82.07 & 97.36 & 94.26 & 97.70 & 98.42&96.39  & 93.75& 86.33&95.54 &90.80 &\bf97.73 \\
5 & 99.46 & 99.64 &   99.82 & 99.10& 99.37& 99.19 & 99.28& 95.87& \bf100.00& 100.00&99.10  \\
6 &  97.92 & 97.59 &   43.12 &83.18& 75.17&83.18  & 90.75& 95.54&98.08 & 90.66&\bf99.30  \\
7 & 88.07 & 82.47 &   90.96& 83.08 & 88.48& 83.08 & 87.56& 98.27& 99.18&98.78 &\bf99.90 \\
8 & 88.14 &  97.62 &77.42  & 89.63 & 87.25&89.63  & 95.81& 98.01& 94.41&93.10 &\bf98.93  \\
9 & 99.87 & 95.60 &87.27  & 96.48 &99.11 &96.48   & 94.21& 96.86&98.62 &\bf99.25 &98.36  \\
\hline \hline
\gr OA (\%) & 75.50 & 86.05 & 79.79 & 77.13 & 73.60&  76.99 & 91.07& 84.65& 85.72& 86.39&  \bf97.81\\
\gr AA (\%) & 86.26 & 88.99 &  85.07& 84.29 &82.91&  80.22 & 90.20& 91.32& 92.85&91.11 & \bf 97.14\\
\gr $\kappa$ & 0.6948 & 0.8187 & 0.7367 &0.7101  & 0.6677&0.7010   & 0.8805&0.8162 & 0.7941& 0.8220& \bf0.9705 \\
\bottomrule[1.5pt]
\end{tabular}}
\label{PU-result}
\end{table*}

To address this, we here propose a spatial-spectral mixture gate to dynamically merge the above features for each position, wherein each location is assigned a group of learnable weights $\tilde{\mathbf{M}} \in \mathbb{R}^{P \times  P\times  2}$ to determine the ratio of various features as follows:
\begin{equation}
\begin{aligned}
&\tilde{\mathbf{M}}_0 = \frac{\exp(\mathcal H(\mathbf{Y}; {\Theta}_{g}))}{\exp(\mathcal H(\mathbf{Y}; {\Theta}_{g}))+\exp(\mathcal H(\mathbf{P}; {\Theta}_{g}))}\\
&\tilde{\mathbf{M}}_1 = \frac{\exp(\mathcal H(\mathbf{P}; {\Theta}_{g}))}{\exp(\mathcal H(\mathbf{Y}; {\Theta}_{g}))+\exp(\mathcal H(\mathbf{P}; {\Theta}_{g}))}\\
\end{aligned},
\end{equation}
where $\mathcal H(*;{\Theta}_{g})$ indicates the feature encoder in SMG. It consists of two fully connected layers with a gaussian error linear unit activation function. Next, a softmax activation function is applied to transfer them into probability maps, whose values are between 0 and 1. Subsequently, we merge the above features as follows:
\begin{equation}
\mathbf{F}=(\tilde{\mathbf{M}}_{0} \cdot  \mathbbm{1}(\tilde{\mathbf{M}}_{0}>\tau )) \odot \mathbf{Y} +(\tilde{\mathbf{M}}_{1}  \cdot  \mathbbm{1}(\tilde{\mathbf{M}}_{1}>\tau ))\odot \mathbf{P},
\end{equation}
where $\tau$ is the threshold for pruning those low contribution features. By employing this gating mechanism, redundant features that do not contribute to HSI classification can be effectively truncated, facilitating the integration of spatial-spectral attributes. The intuition behind SMG is to encourage competition between the two types of features to select the most discriminative ones under different scenarios, thereby achieving a more satisfactory fusion.
\begin{figure}[t]
\scriptsize
  \centering
  \begin{overpic}[width=\linewidth]{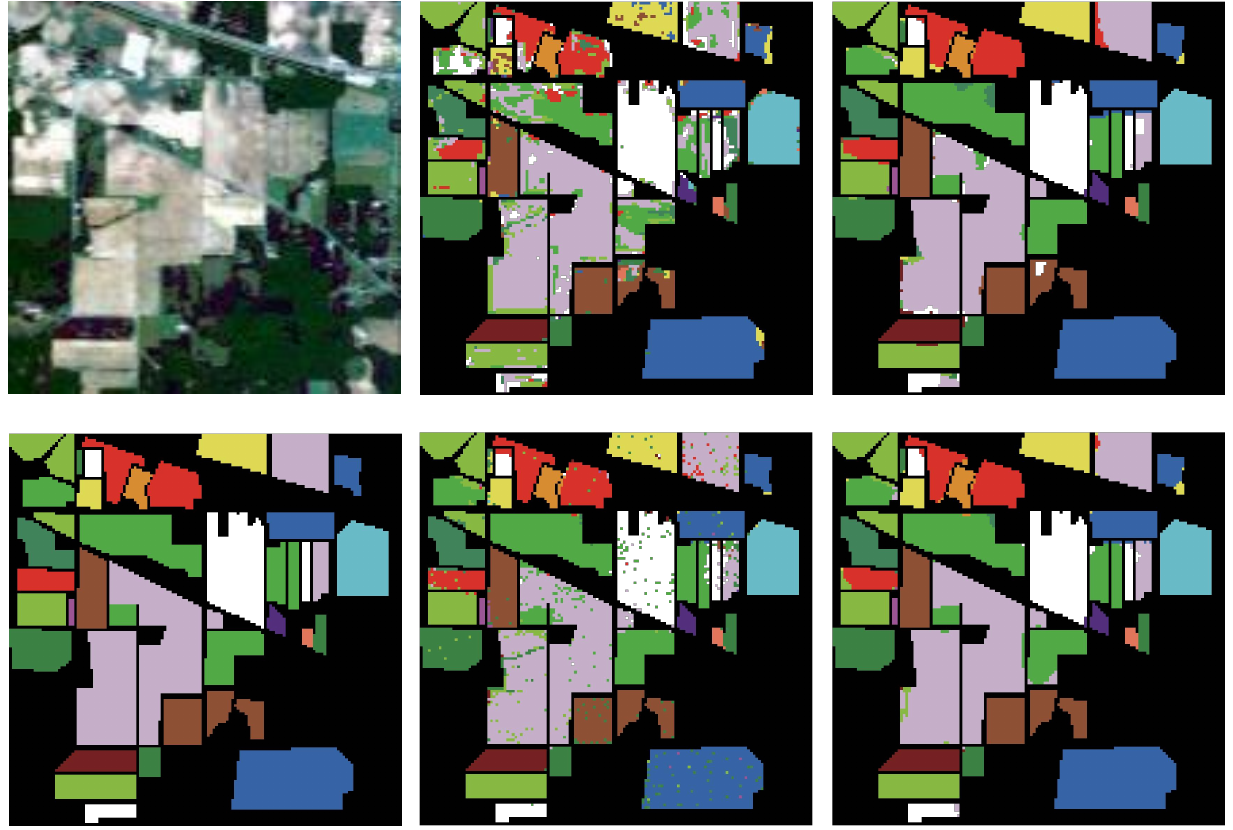}
  \put(7,32.8){{\bf{False-color Map}}}        
  \put(7.7,-2.5){{\bf{Ground Truth}}}    
  \put(38,32.8){{\bf{SpectralFormer\cite{Spectralformer}}}}
  \put(76,32.8){{\bf{SSFTT\cite{SSFTT}}}}
  \put(41,-2.5){{\bf{GraphGST\cite{GraphGST}}}}
  \put(75,-2.5){{\bf{S$^2$Mamba}}}
  \end{overpic}
  % \vspace{-15pt}
  \caption{Qualitative classification results on the Indian Pines dataset.} 
  \label{fig:IP}
  % \vspace{-8pt}
\end{figure}
\begin{figure}[t]
\scriptsize
  \centering
  \begin{overpic}[width=.9\linewidth]{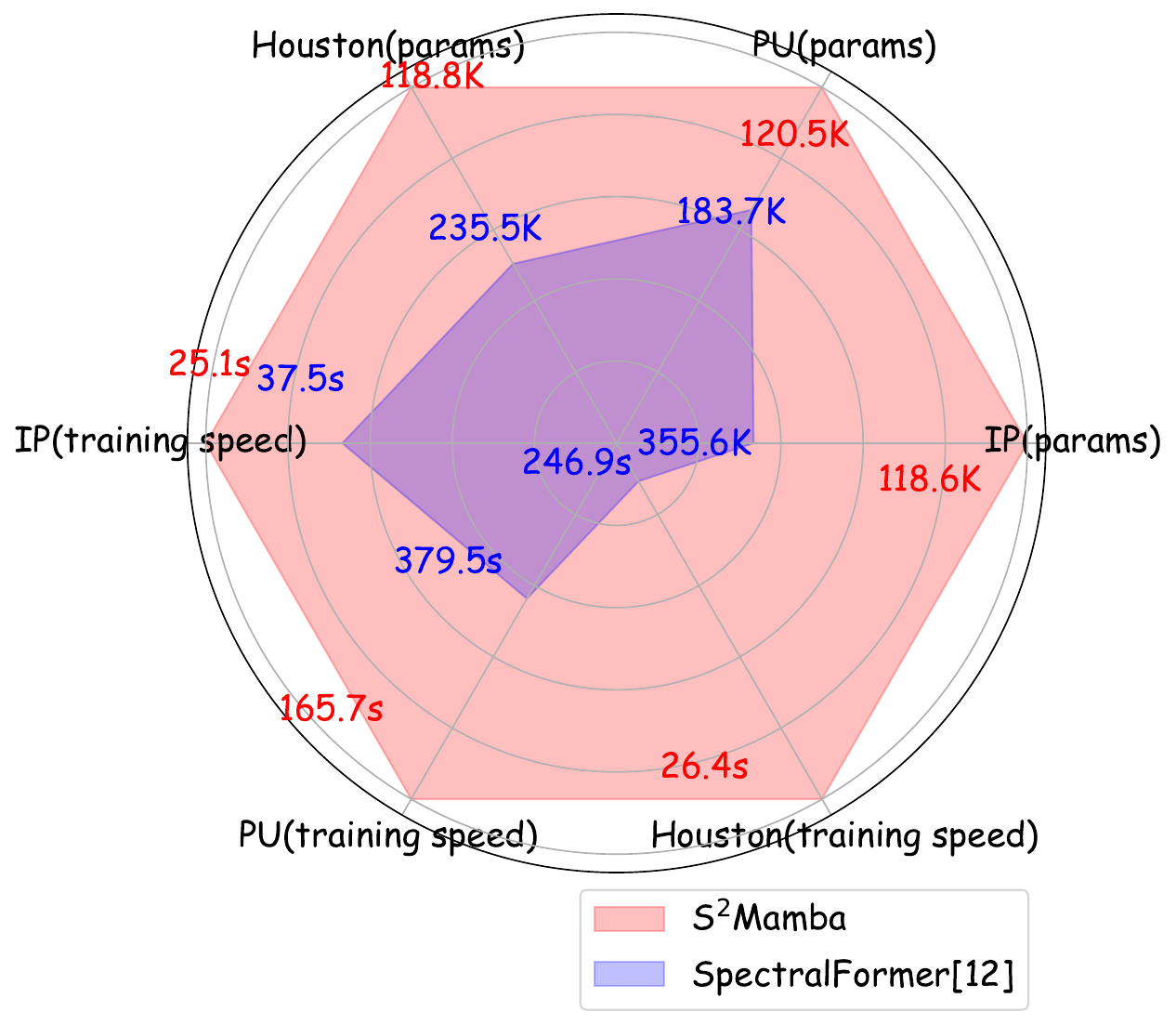}
  \end{overpic}
  % \vspace{-15pt}
  \caption{Comparison of model parameters and running times on the Indian Pines (IP), Pavia University (PU), and Houston 2013 (Houston) datasets.} 
  \label{fig:radar}
  % \vspace{-8pt}
\end{figure}
\subsubsection{\textbf{Computational Complexity Analysis}}
Given the HSI patch data with $K$ spectral bands and $N$ pixels, the proposed S$^2$Mamba encloses four spatial scanning routes for pixels and two spectral scanning routes for spectral bands, respectively. The overall computational complexity can be expressed as follows:
\begin{equation}
\begin{aligned}
\mathcal{O}\left(\text{S$^2$Mamba}\right) &= \mathcal{O}\left(PCS\right) + \mathcal{O}\left(BSS\right) \\
&= \mathcal{O}\left(4N+2K\right)
\end{aligned}
\end{equation}

Since the number of pixels and spectral bands in each HSI patch are considerably larger than 4 and 2, respectively, the computational complexity of the proposed S$^2$Mamba is slighter than Transformer-based methods, whose computational complexity is $\mathcal{O}\left(N^2+K^2\right)$, demonstrating its superiority.

\begin{table*}[t]
\centering
\caption{Comparison with other state-of-the-art methods in terms of (OA\%/AA\%/$\kappa$) on the Houston 2013 test set. The best is in bold.}
% \vspace{-4pt}
\resizebox{0.99\textwidth}{!}{
\begin{tabular}{c||ccc|cc|ccccc|c}
\toprule[1.5pt] \multirow{2}{*}{Category} & \multicolumn{3}{c|}{CNNs} & \multicolumn{2}{c|}{RNN} & \multicolumn{5}{c|}{Transformers} & \multicolumn{1}{c}{Mamba}\\
\cline{2-12}  
% \specialrule{0em}{1pt}{1pt}
& 1-D CNN & 2-D CNN & miniGCN & RNN & CasRNN& ViT & SpectralFormer & morphFormer & SSFTT& GraphGST & \bf S$^2$Mamba\\
\hline \hline
1 & 87.27 & 85.09 & \bf98.39 & 82.34  & 82.62&82.81  & 81.86& 93.63&85.37 &85.66 & 83.10 \\
2 & 98.21 & 99.91 &  92.11 & 94.27& 96.90 &96.62 & 100.00& 98.02&99.81 &99.91 &\bf 100.00  \\
3 &  \bf100.00 & 77.23 &99.60 & 99.60  & 99.60 &99.80 & 95.25& 99.20&95.05 &95.64 &99.60  \\
4 & 92.99 & 97.73 & 96.78 & 97.54 &  96.97&  \bf99.24& 96.12& 95.64& 90.24&91.29 &98.20  \\
5 & 97.35 &  99.53 &97.73 & 93.28  & 97.35 & 97.73&  99.53& 98.95& 100.00& 100.00&\bf100.00  \\
6 & 95.10 & 92.31 &95.10&  95.10  & 95.10 & 95.10& 94.41& 100.00&100.00 &\bf100.00 &95.80  \\
7 & 77.33 &  \bf92.16 & 57.28& 83.77  & 76.21 & 76.77& 83.12& 86.66& 75.00&76.96 & 89.37\\
8 & 51.38 &  79.39 & 68.09 & 56.03 & 44.63&  55.65& 76.73& 83.85& 89.74& \bf89.84& 88.60 \\
9 & 27.95 &  86.31 & 53.92&  72.14 & 64.97 & 67.42& 79.32& 75.92& 76.86&83.57 &\bf92.45 \\
10 &  90.83 & 43.73 & 77.41& 84.17  & 78.28 &68.05 & 78.86& 66.50& 90.92& 91.79&\bf92.57 \\
11 & 79.32 & 87.00 & 84.91 &82.83  & 88.43 & 82.35&  88.71& 82.06& 87.95&91.46 & \bf91.56 \\
12 & 76.56 & 66.28 &77.23  & 70.61 &66.38 & 58.50 &  87.32& 87.22& \bf92.02&90.49 & 90.97 \\
13 & 69.47 &  \bf90.18 & 50.88 &69.12  & 70.53 &60.00 & 72.63& 84.91& 80.00&81.40 &89.12  \\
14 & 99.19 & 90.69 &  98.38 & 98.79& 100.00 &98.79 &  100.00& 100.00& 100.00&100.00 &\bf100.00  \\
15 & 98.10 & 77.80 & 98.52 & 95.98 & 96.62 &98.73 &  99.79& 99.57&100.00 &100.00 &\bf 100.00 \\
\hline \hline
\gr OA (\%) & 80.04 & 83.72 &  81.71& 83.23 & 81.22 &80.41 & 88.01& 88.27&89.62 &90.80 & \bf 93.36\\
\gr AA (\%) & 82.74 & 84.35 & 83.09 & 85.04 & 83.64 &82.50 & 88.91& 90.15&90.87 &91.93 &\bf 94.09 \\
\gr $\kappa$ & 0.7835 & 0.8231 &  0.8018& 0.8183 &0.7967 & 0.7876  & 0.8699& 0.8727&0.8873 &0.8853 &\bf 0.9279 \\
\bottomrule[1.5pt]
\end{tabular}}
\label{Houston-result}
% \vspace{-8pt}
\end{table*}
\begin{figure*}[t]
\small
  \centering
  \begin{overpic}[width=.98\linewidth]{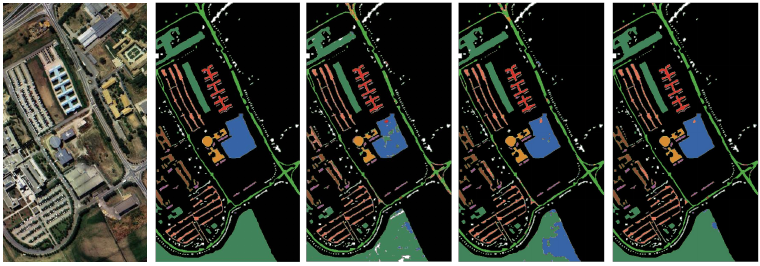}
  \put(2,-1.3){{\bf{False-color Map}}}        
  \put(25,-1.3){{\bf{Ground Truth}}}    
  \put(43,-1.3){{\bf{SpectralFormer\cite{Spectralformer}}}}
  \put(66,-1.3){{\bf{SSFTT\cite{SSFTT}}}}
  \put(86,-1.3){{\bf{S$^2$Mamba}}}
  \end{overpic}
  \caption{Qualitative classification results on the Pavia University dataset.} 
  \label{fig:PU}
\end{figure*}

\section{Datasets and Evaluation Metrics}
In this section, we will present the datasets and evaluation criteria employed in the experiments. 
\subsection{Datasets} 
We conduct evaluations of our S$^2$Mamba on three publicly available datasets, focusing on hyperspectral image classification: Indian Pines, Pavia University, and Houston 2013 datasets. All experiments use the same training and validate samples as \cite{Spectralformer} for a fair comparison, i.e., using a disjoint sampling strategy.

\subsubsection{Indian Pines}
The Indian Pines dataset consists of 145$\times$145 pixels at a ground sampling distance of 20 m and 220 spectral bands spanning the wavelength range of 400–2500 nm, in which 200 spectral bands are preserved after removing 20 noisy and water absorption bands. The dataset is annotated using 16 land cover categories, including crops, trees, and other vegetation. The details are exhibited in Table \ref{data1}. 

\subsubsection{Pavia University}
The Pavia University dataset consists of 610$\times$340 pixels at a ground sampling distance of 1.3 m and 103 spectral bands spanning the wavelength range of 430 to 860 nm, which is annotated using 9 land cover categories, including asphalt, meadows, gravel, trees, metal sheets, bare soil, bitumen, bricks, and shadows. The category details are exhibited in Table \ref{data2}.

\subsubsection{Houston 2013}
The Houston 2013 dataset consists of 349$\times$1905 pixels at a ground sampling distance of 2.5 m and 144 spectral bands spanning the wavelength range of 380 to 1050 nm. The dataset is annotated using 15 land cover categories, including healthy grass, stressed grass, synthetic grass, trees, soil, water, residential, and others. The details are exhibited in Table \ref{data3}.

\begin{figure*}[t]
\small
  \centering
  \begin{overpic}[width=.95\linewidth]{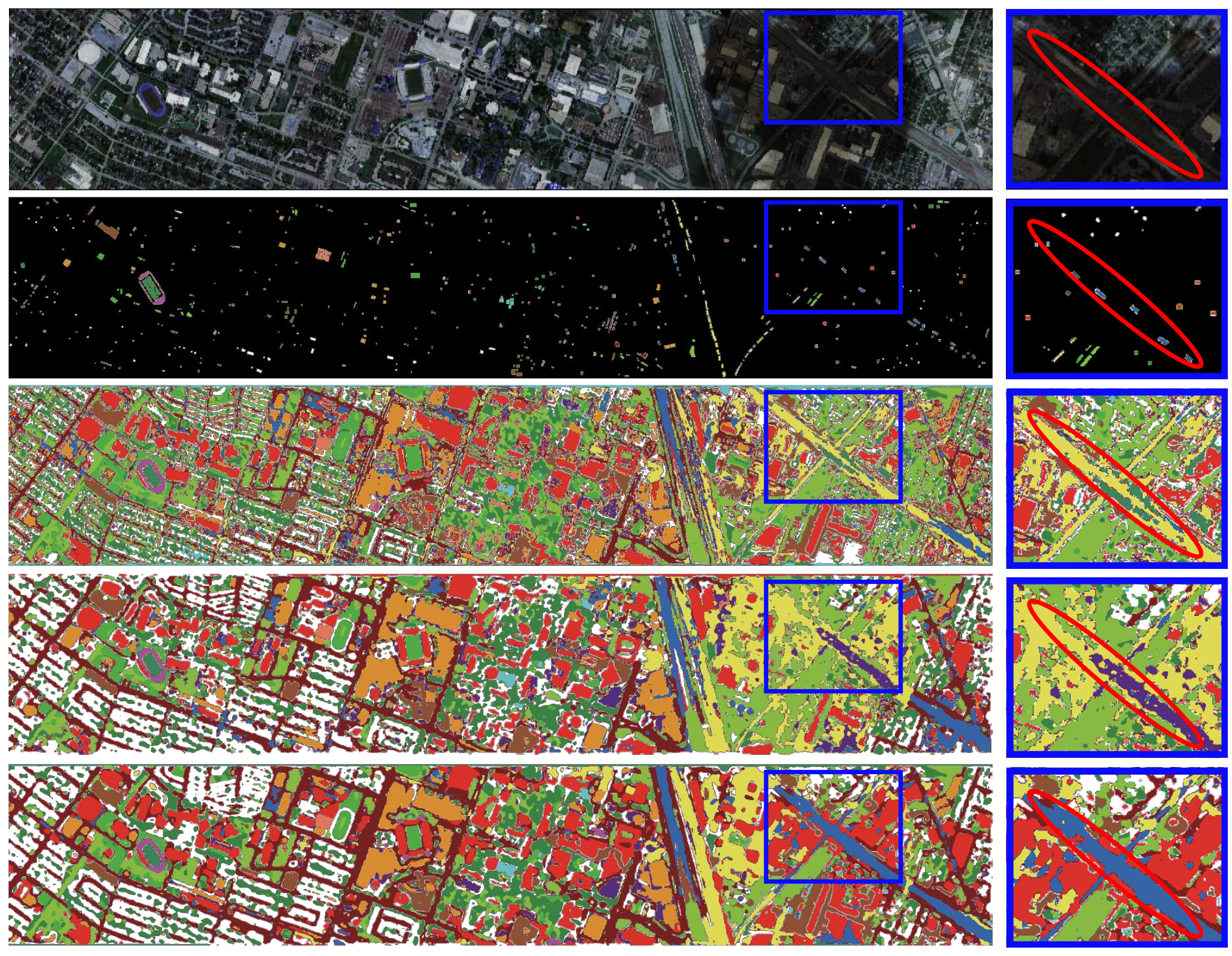}
  \put(100.2,76){\rotatebox{270}{\bf{False-color Map}}}        
  \put(100.2,60){\rotatebox{270}{\bf{Ground Truth}}}    
  \put(100.2,46){\rotatebox{270}{\bf{SpectralFormer\cite{Spectralformer}}}}
  \put(100.2,27.3){\rotatebox{270}{\bf{SSFTT\cite{SSFTT}}}}
  \put(100.2,12.5){\rotatebox{270}{\bf{S$^2$Mamba}}}
  \end{overpic}
  % \vspace{-15pt}
  \caption{Qualitative classification results on the Houston 2013 dataset.} 
  \label{fig:Houston}
  % \vspace{-8pt}
\end{figure*}
% , which is also detailed in the supplementary materials.

\subsection{Evaluation Metrics} 
In this paper, we utilized three commonly used evaluation metrics: overall accuracy (OA), average accuracy (AA), and kappa coefficient ($\kappa$) to evaluate the classification performance of our S$^2$Mamba. 

\section{Experiment}
\subsection{Implementation Details} 
In this study, all the experiments are conducted within the PyTorch framework, utilizing a single NVIDIA GeForce RTX 4090 with 24-GB GPU memory. For the initialization of our S$^2$Mamba, its parameters are randomly initialized by a zero-mean normal distribution with a standard deviation of 0.01. AdamW \cite{adamw} is adopted as the optimizer for S$^2$Mamba, where the exponential learning rate strategy is applied with an initial value of 0.0001. The model is trained with 400 epochs, where the batch size is set to 64. The patch size of inputs is set to 7, 11, and 9 for Indian Pines, Pavia University and Houston 2013, respectively. The number of blocks $N$ is set to 1. The threshold $\tau$ is set to 0.1.

\subsection{Comparison with State-of-the-Arts}
We conduct a comprehensive evaluation of our S$^2$Mamba and compare it with several comparative approaches, including CNN-based (1-D CNN, 2-D CNN, miniGCN \textit{(TGRS'2020)} \cite{GNN_2}), RNN-based (RNN, CasRNN \textit{(TGRS'2019)} \cite{RNN_3}), and Transformer-based (ViT \cite{Vit}, Spectralformer \textit{(TGRS'2021)} \cite{Spectralformer}, SSFTT \textit{(TGRS'2022)} \cite{SSFTT}, morphformer \textit{(TGRS'2023)} \cite{morphformer}, GraphGST \textit{(TGRS'2024)} \cite{GraphGST}) methods, where 1-D CNN, 2-D CNN, RNN, and ViT are implemented following \cite{Spectralformer}. All methods are tested using the optimal experimental settings reported in their papers or re-implemented by their official code.

\subsubsection{Indian Pines} In Table \ref{IP-result}, we benchmark performances on the Indian Pines dataset. The results, presented in Table \ref{IP-result}, demonstrate that our approach significantly outperforms state-of-the-art hyperspectral image classification methods on 14 categories, achieving the best comprehensive performance in terms of OA (97.92\% vs. 97.06\%), AA (98.88\% vs. 98.39\%), and $\kappa$ (0.9761 vs. 0.9664). In particular, we surpass the performance of the typical Transformer-based method, i.e., SpectralFormer \cite{Spectralformer}, which utilizes transformer architecture to extract long-range dependencies from continuous spectral bands. In contrast, our S$^2$Mamba, involving more efficient basic structures and elaborate designs, achieves superior results in terms of OA, AA, and $\kappa$ (e.g., improving the OA from 81.76\% to 97.92\%). Furthermore, Fig. \ref{fig:IP} shows examples of prediction maps. It demonstrates that our S$^2$Mamba is capable of producing accurate predictions of each category.

\begin{table*}[t]
    \centering
    \caption{Ablation study on the impact of each component in S$^2$Mamba. "PCS": Patch Cross Scanning. "BSS": Bi-directional Spectral Scanning. "SMG": Spatial-spectral Mixture Gate.} 
    % \vspace{-4pt}
    \resizebox{0.92\textwidth}{!}{\begin{tabular}{cccccc|ccc|ccc}
    \toprule[1.5pt]
    \multirow[c]{2}{*}{ PCS } & \multirow[c]{2}{*}{ BSS } & \multirow[c]{2}{*}{ SMG } & \multicolumn{3}{c}{Indian Pines} & \multicolumn{3}{c}{Pavia University} & \multicolumn{3}{c}{Houston 2013}  \\
    \cline {4-12}
    \specialrule{0em}{1pt}{1pt}	
    & & & OA (\%) & AA (\%)& $\kappa$& OA (\%) & AA (\%)& $\kappa$& OA (\%) &AA (\%)& $\kappa$ \\
    % \midrule
    \hline \hline
    \specialrule{0em}{1pt}{1pt}
    $\checkmark$& & &96.45&97.99&0.9593&96.42& 96.53&0.9521&90.78&92.05&0.9002 \\
    &$\checkmark$  &  &96.51&98.27& 0.9599&96.24&96.21&0.9493&92.19&93.29&0.9154 \\
    $\checkmark$ &$\checkmark$ & &97.19 &98.56&0.9678&97.17&97.12&0.9620&92.74&93.36&0.9211 \\
    \gr $\checkmark$ & $\checkmark$ & $\checkmark$&\bf 97.92&\bf98.88&\bf0.9761 &\bf 97.81&\bf 97.14&\bf 0.9705&\bf 93.36&\bf 94.09&\bf 0.9279 \\
    \bottomrule[1.5pt]
    \end{tabular}}
\label{ablation-result}
    % \vspace{-8pt}
\end{table*}
\begin{figure*}[t]
	\centering
        % \vspace{-8pt}
	\begin{overpic}[width=0.98\linewidth]{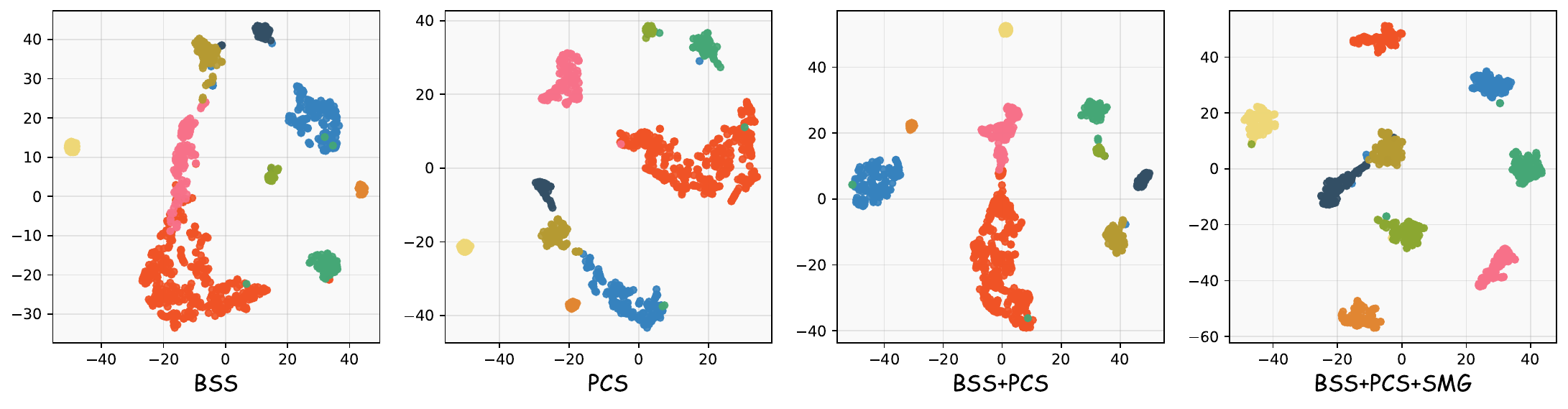}
	\end{overpic}
        % \vspace{-4pt}
	\caption{Feature visualization of different components in S$^2$Mamba using t-SNE \cite{tsne} on the test set of the Pavia University dataset.}
    \label{fig_tsne}
    % \vspace{-18pt}
\end{figure*}

\subsubsection{Pavia University} We further evaluate our S$^2$Mamba on the Pavia University dataset, whose scenario includes numerous complex spatial textures. As illustrated in Table \ref{PU-result}, our method significantly outperforms the comparative techniques with at least 6.74\%, 4.29\%, and 9.00\% in terms of OA, AA, and $\kappa$, respectively. This confirms that our S$^2$Mamba can offer a more satisfactory solution under complex scenarios. In particular, our S$^2$Mamba outperforms it with 6.74\%, 6.94\%, and 9.00\% in terms of OA, AA, and $\kappa$, respectively, compared to the typical Transformer-based method \cite{Spectralformer}. In comparison with the most recent method, i.e. GraphGST \cite{GraphGST}, our method achieves an 11.42\% gain in terms of OA, which demonstrates the effectiveness of our proposed S$^2$Mamba.

Fig. \ref{fig:PU} shows that our approach achieves superior performance than others with more completed prediction maps, such as asphalt (masked in light green color) and meadow (masked in dark green color).

\subsubsection{Houston 2013} Furthermore, we compare our proposed method with the state-of-the-art methods on the more challenging dataset, i.e., Houston 2013. The experimental results, presented in Table \ref{Houston-result}, demonstrate that our S$^2$Mamba significantly outperforms state-of-the-art hyperspectral image classification methods on 7 categories, achieving the best comprehensive performance in terms of OA, AA, and Kappa coefficient. Specifically, it exhibits 2.56\%, 2.16\%, and 4.26\% gains in comparison with the advanced Transformer-based method \cite{GraphGST} on OA, AA, and $\kappa$, respectively. Fig. \ref{fig:Houston} shows classification predictions, which demonstrates that our S$^2$Mamba realizes the most precise results. For instance, our method is capable of accurately identifying highways (masked in blue color) under shadows, demonstrating its effectiveness.

In addition, to verify the efficiency of the proposed S$^2$Mamba, we carried out a comparison of model parameters and running times on the three datasets. For a fair comparison, all experiments are conducted in the same environment. As shown in Fig. \ref{fig:radar}, in comparison with the typical Transformer-based HSI classification model \cite{Spectralformer}, our method can greatly reduce the model parameters and seriously raise the training speed. This confirms the efficiency and lightweight of S$^2$Mamba, which is consistent with our motivation.

\subsection{Ablation Study}
To demonstrate the effectiveness of each component in S$^2$Mamba, we perform a thorough ablation study on Indian Pines, Pavia University, and Houston 2013 datasets. Identical training strategies are employed in all the experiments. 

\begin{figure*}[t]
\scriptsize
\begin{minipage}[t]{0.245\textwidth}
  \centering
  \begin{overpic}[width=\linewidth]{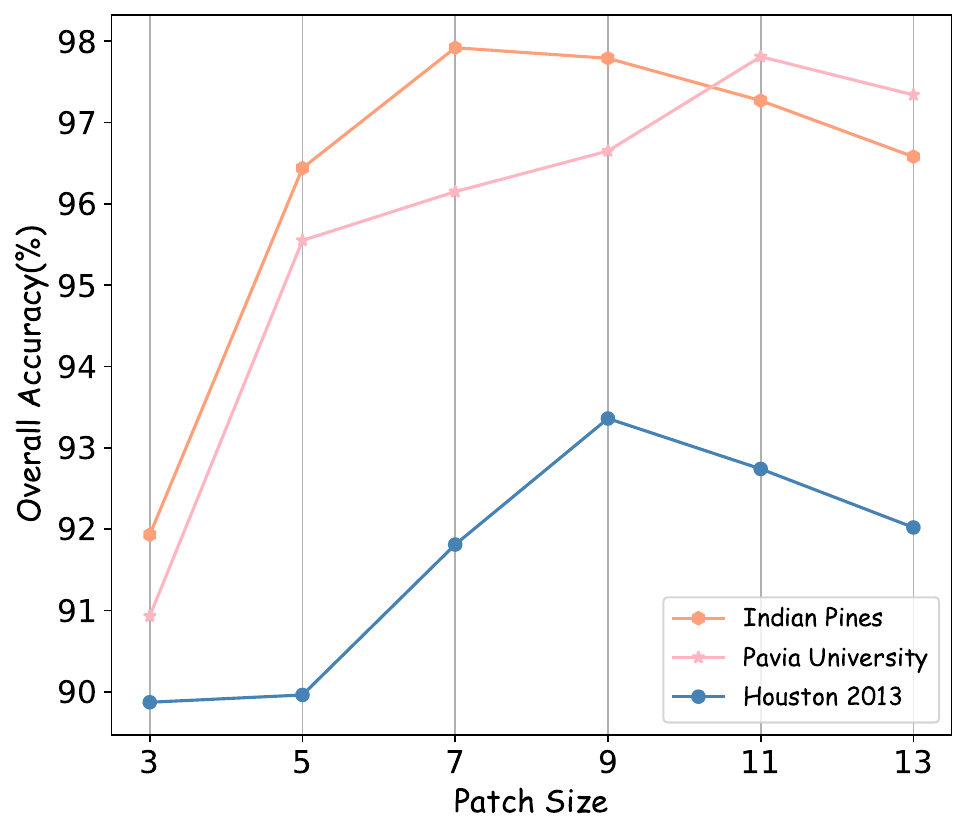}    
  \end{overpic}
  % \vspace{-15pt}
  \caption{Parameter analysis of different patch sizes.} 
  \label{fig:patchsize}
\end{minipage}
% \hfill
\begin{minipage}[t]{0.245\textwidth}
  \centering
  \begin{overpic}[width=\linewidth]{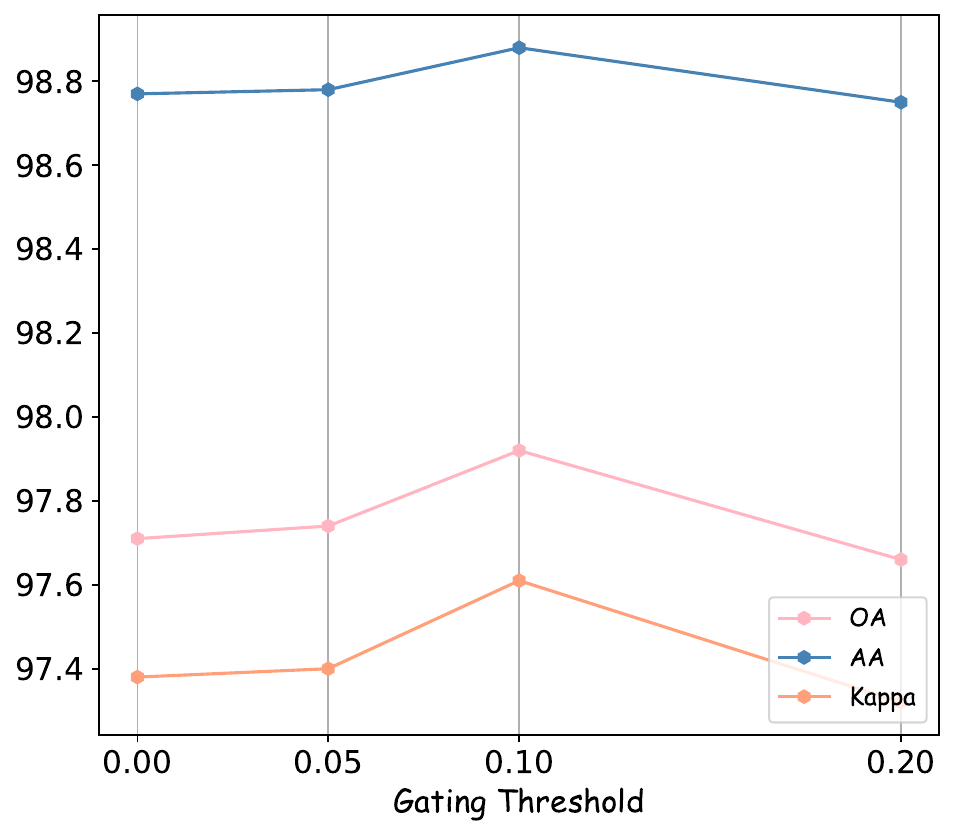}    
  \end{overpic}
  % \vspace{-15pt}
  \caption{Parameter analysis of different gating thresholds.} 
  \label{gating}
\end{minipage}
% \hfill
\begin{minipage}[t]{0.245\textwidth}
  \centering
  \begin{overpic}[width=\linewidth]{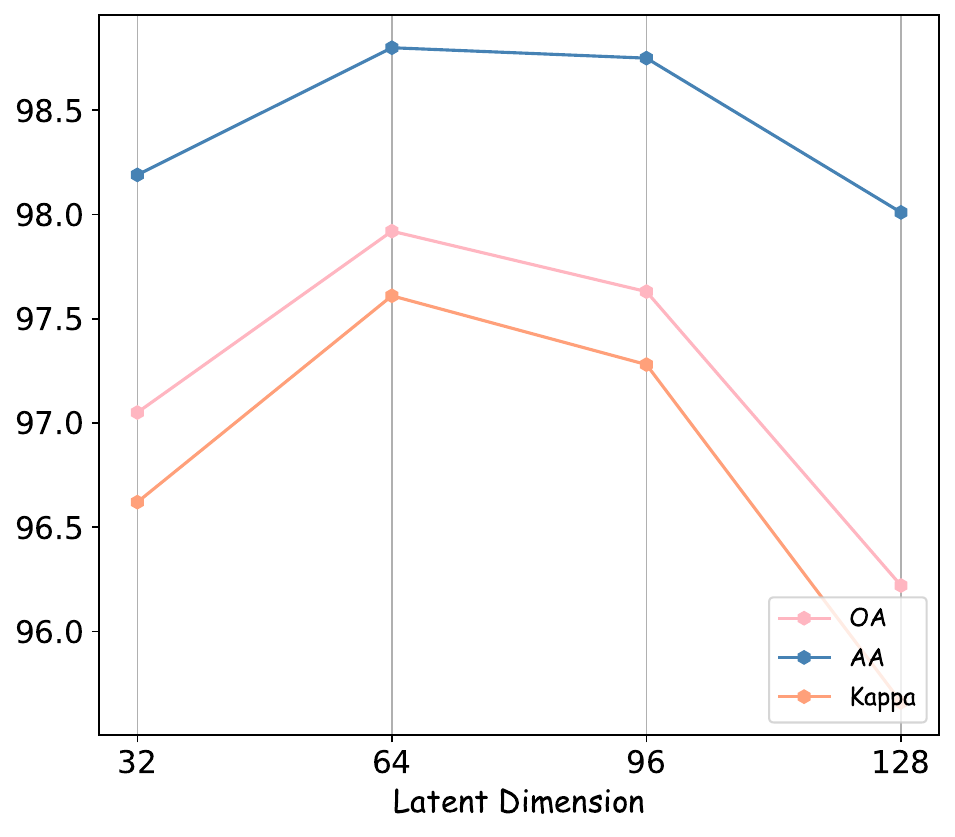}    
  \end{overpic}
  % \vspace{-15pt}
  \caption{Parameter analysis of different latent dimensions.} 
  \label{latent}
\end{minipage}
% \hfill
\begin{minipage}[t]{0.245\textwidth}
  \centering
  \begin{overpic}[width=\linewidth]{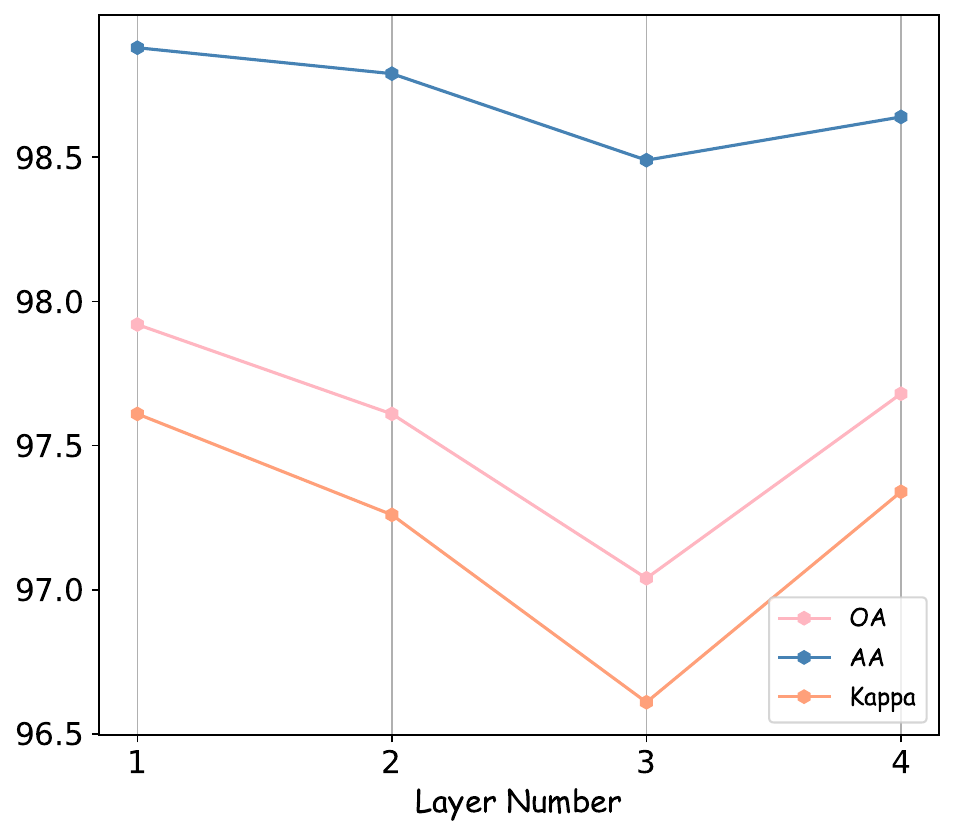}    
  \end{overpic}
  % \vspace{-15pt}
  \caption{Parameter analysis of different layer numbers.} 
  \label{layers}
\end{minipage}
\end{figure*}

% \subsubsection{The effectiveness of each component in S$^2$Mamba.}
Table \ref{ablation-result} shows the impact of each component in S$^2$Mamba. The first row shows the performance using the PCS mechanism, which efficiently considers the spatial relations of pixels by a Mamba-based module. As can be seen, it outperforms most comparative methods with 96.45\%, 96.42\%, and 90.78\% OA on Indian Pines, Pavia University, and Houston 2013 datasets, respectively, due to its effective ability in modeling spatial contexts. Then, by incorporating a bi-directional spectral scanning mechanism, we improve performance to 96.72\%, 97.17\%, and 92.74\% OA on the three datasets. This suggests that the BSS can offer more discriminative cues by scanning continuous spectral bands. Finally, by combining both the PCS, BSS, and SMG, we further boost the performance to 97.92\%/98.80\%/0.9746, 97.81\%/97.14\%/0.9705, and 93.36\%/94.09\%/0.9279 in terms of OA/AA/k on the three datasets, which confirms the SMG module can preferably merge spatial and spectral features for each location by a feature competition. The comprehensive results demonstrate that our S$^2$Mamba can effectively learn discriminative representations of HSI data.

To further analyze the effectiveness of each component in S$^2$Mamba, we visualize the feature distribution under different model settings in Fig. \ref{fig_tsne} using t-SNE \cite{tsne}. It can be seen that each component of S$^2$Mamba can greatly improve the discriminative ability of the classification model, especially for spatial-spectral feature fusion. When jointly applying BSS and PCS modules, the feature disparity of S$^2$Mamba significantly increases. After inserting all the modules, S$^2$Mamba achieves the best performance, demonstrating the significance of each component.

\subsection{Parameter Analysis}
In this section, we conduct comprehensive experiments to analyze the hyperparameters within the proposed S$^2$Mamba. The remaining hyperparameters are consistently set to optimal values across all experiments. All the datasets are employed to demonstrate the robustness of the proposed method on fluctuations of hyperparameters.

% \subsubsection{Parameter analysis of each component in S$^2$Mamba.}
Specifically, we study the effectiveness of hyperparameters in S$^2$Mamba, including patch size, gating threshold, latent dimension, and layer numbers. As shown in Fig. \ref{fig:patchsize}, it can be observed that the optimal patch sizes of Indian Pines, Pavia University, and Houston 2013 are 7, 11, and 9, separately, which is consistent with the expectation that the last two datasets comprise more complex spatial boundaries, thereby larger patch inputs are required. Fig. \ref{gating} demonstrates that firming the gating threshold to 0.1 can achieve satisfactory results, which can filter those redundant features. Figs. \ref{latent} and \ref{layers} denote that 64 and 1 are the optimal values of hidden dimension and layer number, respectively. This confirms our S$^2$Mamba can achieve state-of-the-art performance with a lightweight network, whereas deeper layers or larger hidden dimensions can not yield additional improvements.

\section{Conclusion}
In this paper, we proposed S$^2$Mamba, a novel architecture for hyperspectral image classification. S$^2$Mamba comprises a Patch Cross Scanning mechanism and a Bi-directional Spectral Scanning mechanism for learning the contextual information from spatial and spectral aspects, respectively, which utilize the selective structured state space models as alternatives to self-attention mechanisms for capturing long-range dependency with linear complexity, thereby efficiently improving the results. Furthermore, to optimally merge the above features, a Spatial-spectral Mixture Gate was proposed to adjust the fusion ratio for each position with learnable matrices, further enhancing the classification performance. Finally, we evaluated the proposed S$^2$Mamba on three public hyperspectral image classification datasets, and the experimental results verified its superiority.

Throughout this paper, we investigated a novel hyperspectral image classification architecture based on state space models, while more potential tasks in hyperspectral image analysis remain uncovered. Considering the efficiency and effectiveness of Mamba models, we look forward to generalizing S$^2$Mamba to improve other tasks in HSI in the future.



\bibliographystyle{IEEEtran}
\bibliography{IEEEabrv,mybibfile}
\ifCLASSOPTIONcaptionsoff

\newpage
\fi
\newpage
\newpage
\newpage


\end{document}